\documentclass[letterpaper]{article} 
\usepackage[]{aaai2026}  
\usepackage{times}  
\usepackage{helvet}  
\usepackage{courier}  
\usepackage[hyphens]{url}  
\usepackage{graphicx} 
\usepackage{natbib}  
\usepackage{caption} 
\usepackage{algorithm}
\usepackage{algorithmic}
\usepackage{newfloat}
\usepackage{listings}
\usepackage{amsmath}
\usepackage{tipa}
\usepackage{amsfonts}
\usepackage{multirow} 
\usepackage{graphicx} 
\usepackage{booktabs} 
\usepackage{amssymb}
\usepackage{indentfirst}
\usepackage{cleveref}
\crefname{figure}{Figure}{Figures}
\crefname{table}{Table}{Tables}
\crefname{equation}{Equation}{Equations}
\usepackage{helvet}
\usepackage{courier}
\usepackage{xcolor}
\usepackage{newfloat}
\usepackage{listings}
\DeclareCaptionStyle{ruled}{labelfont=normalfont,labelsep=colon,strut=off} 
\floatstyle{ruled}
\newfloat{listing}{tb}{lst}{}
\floatname{listing}{Listing}
\nocopyright 

\title{M2DAO-Talker: Harmonizing Multi-granular Motion Decoupling \\ and Alternating Optimization for Talking-head Generation}
\author{
    Kui Jiang\textsuperscript{\rm 1}, Shiyu Liu\textsuperscript{\rm 1}, Junjun Jiang\textsuperscript{\rm 1},  Hongxun Yao\textsuperscript{\rm 1}, Xiaopeng~Fan\textsuperscript{\rm 1}
}
\affiliations{
    \textsuperscript{\rm 1}Harbin Institute of Technology \\
    jiangkui@hit.edu.cn, liushiyu\_aiia@stu.hit.edu.cn, 
    jiangjunjun@hit.edu.cn,  h.yao@hit.edu.cn,  fxp@hit.edu.cn
}

\usepackage{bibentry}

\begin{document}

\maketitle

\begin{abstract}
  Audio-driven talking head generation holds significant potential for film production. While existing 3D methods have advanced motion modeling and content synthesis, they often produce rendering artifacts, such as motion blur, temporal jitter, and local penetration, due to limitations in representing stable, fine-grained motion fields. Through systematic analysis, we reformulate talking head generation into a unified framework comprising three steps: video preprocessing, motion representation, and rendering reconstruction. This framework underpins our proposed M2DAO-Talker, which addresses current limitations via multi-granular motion decoupling and alternating optimization. 
  Specifically, we devise a novel 2D portrait preprocessing pipeline to extract frame-wise deformation control conditions (motion region segmentation masks, and camera parameters) to facilitate motion representation. 
  To ameliorate motion modeling, we elaborate a multi-granular motion decoupling strategy, which independently models non-rigid (oral and facial) and rigid (head) motions for improved reconstruction accuracy.
  Meanwhile, a motion consistency constraint is developed to ensure head-torso kinematic consistency, thereby mitigating penetration artifacts caused by motion aliasing. 
  In addition, an alternating optimization strategy is designed to iteratively refine facial and oral motion parameters, enabling more realistic video generation.  
  Experiments across multiple datasets show that M2DAO-Talker achieves state-of-the-art performance, with the 2.43 dB PSNR improvement in generation quality and 0.64 gain in user-evaluated video realness versus TalkingGaussian while with 150 FPS inference speed. 
\end{abstract}

%

\section{Introduction}
The generation of talking head videos~\cite{wang2024high} significantly enhances digital human realism by aligning with human perceptual expectations, making it particularly valuable for film production, immersive education platforms, and so on. 


Traditional studies~\cite{wang2023seeing, zhong2023identity,guan2023stylesync,9557828} employ robust content generation capabilities of generative adversarial networks (GANs) for lip-sync prediction, demonstrating preliminary success in audio-driven synthesis. However, their lack of explicit motion field modeling and geometric priors leads to limited capacity for fine-grained facial dynamics, and artifacts from fixed head pose constraints. 
Recently, the 3D reconstruction area is experiencing vigorous development with deep learning technologies, such as Neural Radiance Fields~\cite{mildenhall2021nerf,yang2025scalable} (NeRF) and 3D Gaussian Splating~\cite{kerbl20233d,fei20243d} (3DGS). 
For talking head generation, NeRF-based methods~\cite{guo2021ad, tang2022real} ameliorate structural stability through tri-plane hash encoders that compress dynamic heads into low-dimensional subspaces. While enhancing identity preservation and texture generation, these methods~\cite{li2023efficient, peng2024synctalk} still face critical limitations. 
For example, they lack precise motion representation due to direct modeling of the relationship between position, color, and density, causing facial distortions during rapid expressions. 

Addressing these issues, 3D Gaussian Splatting (3DGS) offers a promising alternative to NeRF, delivering comparable rendering quality with significantly faster inference. Recent 3DGS variants~\cite{cho2024gaussiantalker,li2025talkinggaussian} enhance temporal stability by learning explicit Gaussian fields to model motion patterns. Some approaches~\cite{li2025talkinggaussian} further acknowledge motion field heterogeneity across facial regions, introducing face-mouth decomposition for fine-grained motion representation.
Despite improved oral rendering, three key challenges persist in 3DGS methods.
First, the extraction and segmentation of motion regions via BiseNet~\cite{yu2018bisenet} has been prone to produce imprecise edges, resulting in blurred edges or piercing phenomena during Gaussian field modeling.
Second, adopting explicit geometric priors via 3D Morphable Models~\cite{paysan20093d} (3DMM) for frame-wise head pose estimation~\cite{li2025talkinggaussian} causes obvious inter-frame head jitter, where the mixed effects of head pose and facial expression on 3DMM are ignored. 
Third, the lack of joint optimization between head and torso movement results in kinematic discontinuity and unnatural visual effect. 
Overall, 3DGS-based technologies~\cite{cho2024gaussiantalker,li2025talkinggaussian} lack consistency and thoroughness when decoupling motion, resulting in unnatural, warped facial repression and piercing phenomena in synthesized videos. 

To circumvent these challenges, we propose M2DAO-Talker--a unified framework integrating multi-granular motion decoupling and alternating optimization for audio-driven head-talking synthesis. 
Our M2DAO-Talker approach systematically organizes the task through three coordinated components: video preprocessing, motion representation, and rendering reconstruction. 
Preprocessing is a key step in achieving precise motion modeling. We implement a high-quality 2D talking portrait video preprocessing pipeline to achieve frame-wise extraction of deformation control parameters, motion region segmentation 
and camera parameters estimation. 
In this way, we can produce the hyperparameters of motion representation, accurate camera parameters, fine-grained semantic masks, and so on. 
These inputs facilitate accurate motion modeling with specific priors and significantly alleviate simulated errors. 

In addition, we have developed a novel multi-granular motion decoupling scheme to bolster motion expression in M2DAO-Talker. 
Specifically, in addition to modeling rigid head rotation through camera parameters, we adopt the common practice of separating non-rigid head motion into facial and oral components using two dedicated branches: a Face Branch for capturing macro-scale expressions (\emph{e.g.}, eyebrow movement, eye blinks) and an Inside Mouth Branch for modeling finer oral articulations such as phoneme-dependent tongue positions.
Meanwhile, we have developed a motion consistency constraint to maintain head-to-torso kinematic consistency. Concretely, the unified representation of torso motion is learned with the Face Branch to guide more reasonable Gaussian field modeling of facial contours, producing visually pleasing and natural reconstruction results. 

To achieve more coherent facial reconstruction, we introduce an alternate optimization strategy that cyclically updates parameters between the Face Branch and the Inside Mouth Branch. 
In this way, it encourages the network to dynamically adjust the convergence path based on the optimization errors of specific branches, enhancing physically consistent blending at facial/oral boundaries.
In general, the contributions are as follows.
\begin{itemize}
\item   
We present a novel M2DAO-Talker approach for audio-driven talking head generation, which ameliorates  stability and consistency of produced video by unifying  multi-granular motion decoupling and alternating optimization. 

\item We elaborate the video preprocessing pipeline to facilitate stable and precise motion field reconstruction, which simultaneously enables frame-wise deformation control extraction, semantic motion region segmentation, and camera parameter estimation. 

\item We introduce the multi-granular motion decoupling scheme, an innovative solution that implements division representation of head, facial, and oral motions, cooperating with a motion consistency constraint and alternate optimization strategy to judiciously mitigate timing jitter and local penetration phenomenon.  

\end{itemize}

\begin{figure*}[!ht]
\flushleft
\centering
\includegraphics[width=0.95\linewidth]{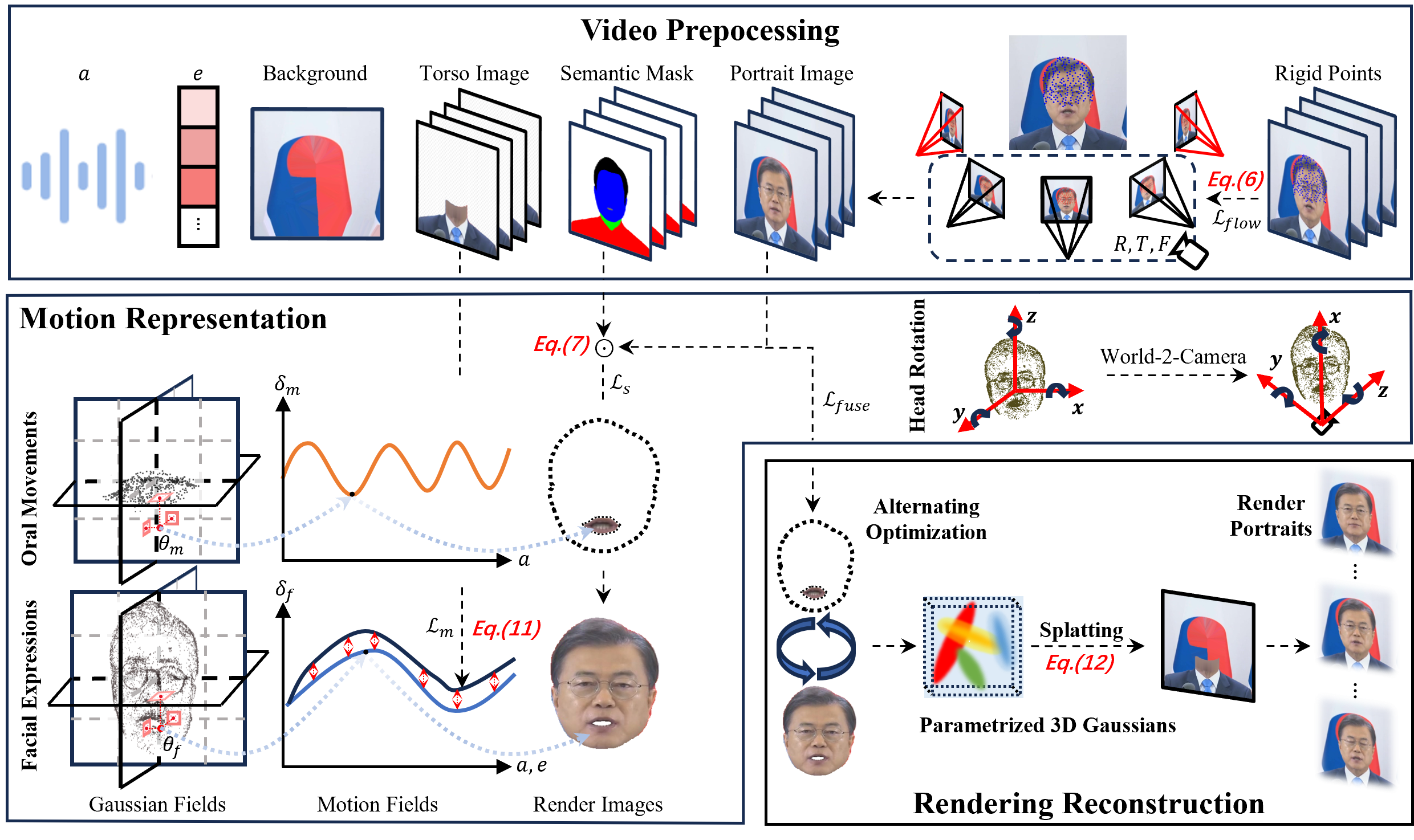}
\vspace{-4mm}
\caption{
The M2DAO-Talker pipeline comprises three stages. i) Video Preprocessing: it extracts key features from input videos, involving portrait image $I$, mouth movement feature $a$, semantic mask $M$, background image $I_{\text{bg}}$, facial expression feature $e$, and camera projection parameters. ii) Motion Representation: head motion is categorized into three components: head rotation, facial expressions, and oral movements. Head rotation is parameterized using a scaling matrix $T$, rotation matrix $R$, and focal length $F$, while facial expressions and oral movements are modeled by two separate motion branches bashed on 3D Gaussian primitives. Region-specific supervision is applied to each branch using $I\odot M$ as a localized training signal, and facial deformations are regularized to maintain coherence with torso motion. iii) Rendering Reconstruction: alternating optimization cyclically updates parameters between the Face Branch and the Inside Mouth Branch, using full portrait $I$ as ground truth. 
}
\label{fig:pipeline}
\vspace{-4mm}
\end{figure*}

\section{Method}
As illustrated in \cref{fig:pipeline}, our proposed framework, \textbf{M2DAO-Talker}, synthesizes audio-driven talking-head videos from monocular portrait footage of a target speaker. 
We propose a multi-granular, fully disentangled motion representation for head motion. Specifically, our 2D video preprocessing pipeline first decouples rigid \textbf{head rotation} and segments distinct facial motion regions. Using semantic masks, non-rigid motion is further decomposed into \textbf{oral movements} and \textbf{facial expressions}. Finally, a motion consistency constraint and an alternating optimization strategy are introduced to enforce coherent dynamics across motion regions. Preliminaries and details of each module  follow in subsequent sections.

\subsection{Preliminaries}

\noindent\textbf{3D Gaussian Splatting.} 
3D Gaussian splatting (3DGS) represents 3D information using anisotropic 3D Gaussians. Given a set of Gaussian primitives $\theta$ and camera parameters, this method computes pixel-level colors  $C$ through differentiable rendering.

Each Gaussian primitive $\theta$ is parameterized by five attributes, involving
the center position $\mu \in \mathbb{R}^3$ (mean of the Gaussian distribution), 
the scaling factor $s \in \mathbb{R}^3$,
the rotation quaternion $q \in \mathbb{R}^4$,
the opacity $\alpha \in \mathbb{R}$, and the $K$-order spherical harmonics coefficients $SH \in \mathbb{R}^{3(k+1)(k+1)}$ for view-dependent color representation. 

The distribution of the Gaussian primitive ($\theta=\{\mu,s,q,\alpha,SH\}$) is defined by its mean $ \mu$ and the covariance matrix $\Sigma \in \mathbb{R}^{3 \times 3}$, depicted as
\begin{equation}
\label{eq:GS1}
g(x)=\exp \left(-\frac{1}{2}(x-\mu)^{T} \Sigma^{-1}(x-\mu)\right),
\end{equation}
where $\Sigma = RSS^T R^T$ . Here, $S=\text{diag}(s)$ is the scaling matrix, and $R$ is the rotation matrix derived from $q$. 

For 2D projection, 3D Gaussians computes the projected covariance $\Sigma'$ using 
\begin{equation}
\label{eq:GS4}
\Sigma' = JW \Sigma W^T J^T,
\end{equation}
where $W$ is the world-to-camera transformation matrix, and $J$ is the Jacobian of perspective projection.
Pixel colors $C$ are rendered via alpha-blending of ordered Gaussians with
\begin{equation}
\label{eq:GS2}
C=\sum_{i=1} c_{i} \alpha_{i}^{\prime} \prod_{j=1}^{i-1}\left(1-\alpha_{j}^{\prime}\right),
\end{equation}
where $c_i$ denotes the view-dependent color from spherical harmonics, and $\alpha' $ represents the projected opacity. The total pixel opacity $A$ is: 

\begin{equation}
\label{eq:GS3}
A=\sum_{i=1} \alpha_{i}^{\prime} \prod_{j=1}^{i-1}\left(1-\alpha_{j}^{\prime}\right).
\end{equation}

\subsection{Video Preprocessing}
\label{sec:preprocessing}
We design a structured preprocessing pipeline to extract semantically  disentangled motion signals from talking portrait videos, providing  a clean and physically grounded foundation for subsequent deformation modeling. Our pipeline focuses on two key aspects: spatial segmentation of motion-sensitive regions and robust estimation of head pose trajectories. Details of audio-visual encoding and facial representations are deferred to the Appendix. 

\noindent\textbf{Motion Region Segmentation (MRS).} 
Facial motion during speech is inherently localized, driven by specific muscular activations rather than uniform deformation. To accurately model such behavior, we introduce a hybrid segmentation framework that refines the delineation of motion-relevant regions with greater granularity and robustness. Rather than relying solely on conventional facial parsing~\cite{yu2018bisenet}, we incorporate spatial priors derived from  position-guided foreground extraction. Specifically, region masks are initialized based on sparse positional prompts (e.g., eyes, mouth, torso), which guide a spatial encoder~\cite{ravi2024sam} to produce coherent foreground boundaries:
\begin{equation}
\label{eq:MR}
M_R = \text{Enc}_p(P_{le}, P_{re}, P_{m}, P_{lt}, P_{rt}),
\end{equation}
where $P_*$ means the key positional prompt extracted from the landmarks.

This enables more stable localization even under challenging conditions such as occlusion or background clutter. The resulting mask is then refined into semantically meaningful subregions $M_R = \{M_{\text{f}}, M_{\text{m}} \}$ through a two-stage parsing strategy. In particular, we enhance the oral region segmentation~\cite{yu2018bisenet} by incoporating a fine-grained teeth-aware parser~\cite{kvanchiani2024easyportraitfaceparsing}, which resolves common edge ambiguities caused by texture homogeneity around the lips and teeth. This segmentation design serves  as the basis for spatially disentangled deformation learning in later stages.


\noindent\textbf{Head Pose Estimation (HPE).} 
To model rigid motion independently of non-rigid deformations, we formulate a motion decomposition strategy~\cite{yao2022dfa} that yields stable camera parameters aligned with physically plausible head movement. Although prior approaches~\cite{li2025talkinggaussian, cho2024gaussiantalker} often regress pose parameters through landmark alignment, they are prone to drift due to entangled expression dynamics.
\begin{figure}[!htp]
    \centering
    \includegraphics[width=\linewidth]{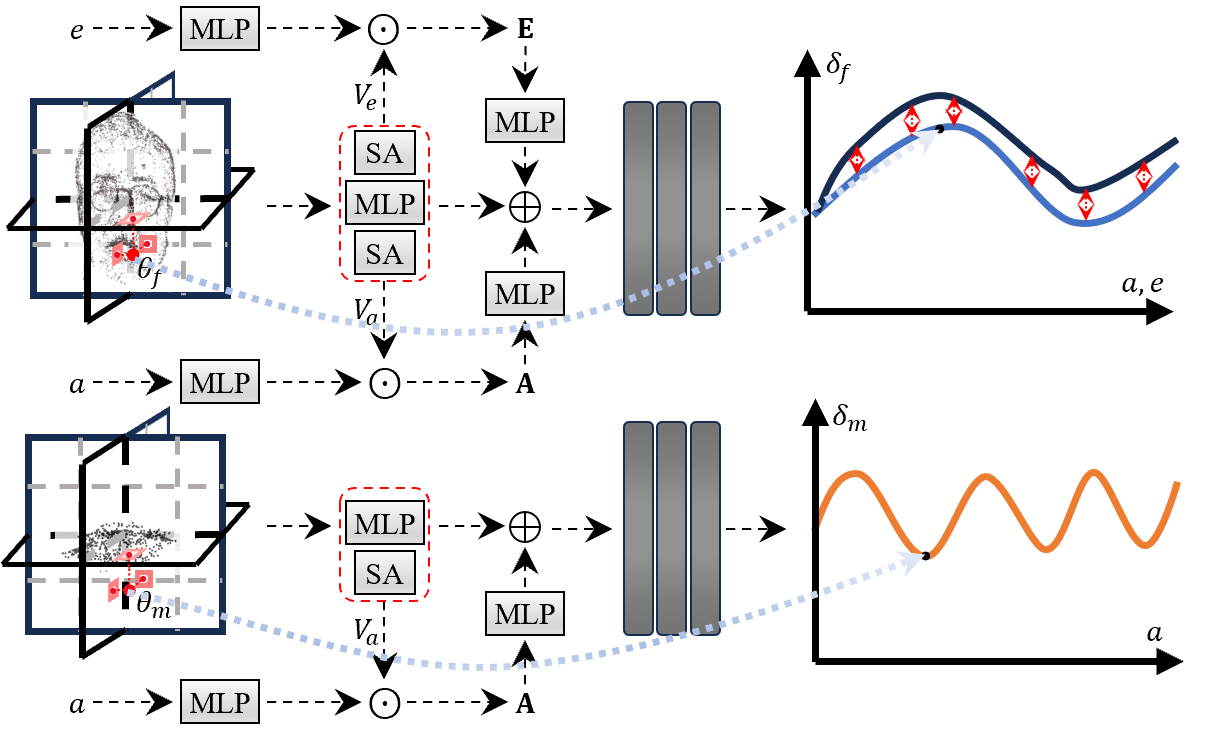}\vspace{-2mm}
    \caption{Detailed architecture of the network designed to capture non-rigid facial dynamics. ``SA'' stands for Self-Attention module.}
    \label{fig:MR}
\vspace{-3mm}
\end{figure}
In contrast, we propose to decouple rigid trajectories by identifying deformation-invariant regions across time. Specifically, we compute per-frame motion vectors and extract temporally stable keypoints (e.g., ears, hairlines) using Laplacian-filtered flow magnitude. These keypoints define a rigid motion trajectory $\{K_t\}_{t=1}^{T}$ that is minimally affected by facial articulation.
The scaling matrix $T$, rotation matrix $R$, and focal length $F$ of camera are then optimized by aligning projected 3D landmarks $P_t$ with the tracked rigid trajectory:
\begin{equation}
\label{eq:HMD}
\mathcal{L}_{\text{flow}} = \sum_{t=1}^{T} \lVert P_t(R_{\text{opt}}, T_{\text{opt}}, F_{\text{opt}}) - K_t \rVert_2.
\end{equation}

To ensure temporal stability, frames with high flow error are filtered based on sequence-level statistics. Importantly, we preserve these frames for alternating optimization by excluding them only from the motion consistency constraint, thereby maintaining expression diversity while suppressing rigid motion noise.

\subsection{Motion Representation}
\label{sec:motion}

Our framework achieves multi-granular motion disentanglement into three distinct components: head rotation, oral movements, and facial expressions. While prior approaches attempt similar factorization, they often exhibit residual entanglement due to insufficient rigid motion modeling or inadequate semantic priors. In contrast, our method explicitly separates these motion components during preprocessing, establishing a robust foundation for downstream learning.

Specifically, we use HPE to extract rigid head rotation and MRS to localize deformable facial subregions. These steps yield high-quality initialization of semantic masks and 3D camera parameters, ensuring physically plausible alignment and precise regional control. With these decoupled signals, we deploy two dedicated deformation branches: a Face Branch for full facial expressions and an Inside Mouth Branch for internal articulations. Region masks ${M_\text{f}, M_\text{m}}$ are applied during both rendering and loss computation to enforce spatial separation and prevent interference between motion types.
\begin{equation}
\label{eq:I_face}
I_{\text{f}} = I \odot M_{\text{f}}, \quad I_{\text{m}} = I \odot M_{\text{m}},
\end{equation}
where $I$ is the input frame and $\odot$ denotes element-wise multiplication. We can further remove $I_\text{m}$ and $I_\text{f}$ from $I$ to obtain the background imgae $I_\text{bg}$ that contains torso motion.

As illustrated in \cref{fig:MR}, we introduce an anatomically grounded attention mechanism to further decouple expressions from speech-induced dynamics. Phoneme-aware features $\mathbf{A} = \text{Enc}(a) \odot V_a$ and expression embeddings $\mathbf{E} = \text{Enc}(e) \odot V_e$ are generated via attention-guided MLPs~\cite{guo2022beyond}, modulating localized Gaussian offsets $\delta = \{\Delta \mu, \Delta s, \Delta q\}$:
\begin{align}
\label{eq:Cface}
\delta_{\text{f}} &= \text{MLP} \left( \mathcal{H}(\mu_{\text{face}}) \oplus \mathbf{A} \oplus \mathbf{E} \right), \\
\label{eq:Cmouth}
\delta_{\text{m}} &= \text{MLP} \left( \mathcal{H}(\mu_{\text{mouth}}) \oplus \mathbf{A} \right),
\end{align}
where $\mathcal{H}: \mathbb{R}^3 \rightarrow \mathbb{R}^{36}$ is a tri-plane hash encoder, and $\oplus$ denotes feature concatenation. 

\noindent\textbf{Motion Consistency Constraint (MCC).}  
To further promote biomechanical coherence, we introduce a motion-level constraint that aligns facial deformations with global torso dynamics. This is implemented through a novel \textit{full-portrait supervision} strategy applied during dynamic refinement. 

Following standard practice for static initialization, both branches first reconstruct region-specific appearance using pixel-wise $\ell_1$ and structural DSSIM losses:
\begin{align}
\label{eq:Ls_face}
\mathcal{L}_{\text{s}}(\hat{I}_{\text{r}}, I_{\text{r}}) &= \ell_1(\hat{I}_{\text{r}}, I_{\text{r}}) + \lambda \mathcal{L}_{\text{SSIM}}(\hat{I}_{\text{r}}, I_{\text{r}}),
\end{align}
where $\hat{I}_{\text{r}}=\{\hat{I}_{\text{f}},\hat{I}_{\text{m}}\}$ denotes predicted  region images and ${I}_{\text{r}}=\{{I}_{\text{f}},{I}_{\text{m}}\}$ denotes ground-truth. 

\textbf{Our core innovation} lies in the dynamic learning, where we introduce full-portrait supervision by compositing facial renderings with background using facial transparency $A_{\text{f}}$: $\hat{I}_{\text{bg}} = \hat{I}_{\text{f}} \times A_{\text{f}} + I_{\text{bg}} \times (1 - A_{\text{f}})$. This enables global optimization through combined objectives: 
\begin{equation}
\label{eq:Lm_face}
\mathcal{L}_{\text{m}} = \mathcal{L}_{\text{s}}(\hat{I}_{\text{bg}}, I) + \gamma \mathcal{L}_{\text{LPIPS}}(\hat{I}_{\text{bg}}, I).
\end{equation}
Unlike prior work constrained to facial regions, this novel supervision strategy explicitly couples facial motion with torso dynamics. However, for the Inside Mouth Branch ,we still adopt specific region constraints for optimization.

\begin{figure}[]
    \centering
    \includegraphics[width=\linewidth]{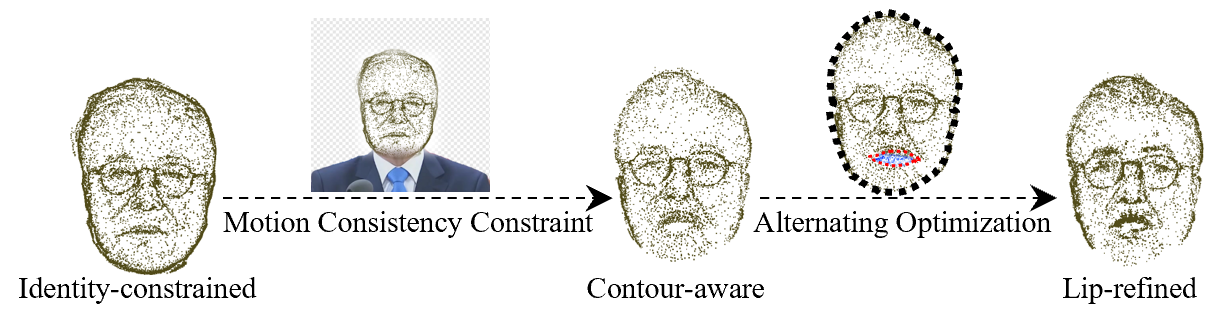}\vspace{-2mm}
    \caption{Iterative optimization of the facial 3DGS point cloud. }
    \label{fig:MS}
\vspace{-3mm}
\end{figure}

This formulation enforces harmonious deformation between facial regions (especially the audio-responsive lower face) and adjacent torso motion. By simultaneously optimizing the transparency parameter $\alpha$ for each 3D Gaussian, we generate temporally consistent alpha mattes that preserve boundary details during face-torso transitions. As shown in \cref{fig:MS}, this constraint sharpens point cloud contours along face-torso interface, improving motion stability and anatomical coherence without architectural overhead.

\subsection{Alternating Optimization Reconstruction}
\label{sec:altopt}

Synthesizing photorealistic talking heads from disentangled motion sources requires coordinated facial and intra-oral rendering. However, joint optimization under motion consistency constraints often introduces boundary artifacts like lip color leakage and tooth discoloration. We attribute these to imbalanced supervision: i) the facial mask $M_\text{f}$ typically excludes inner-lip regions; ii) under-constrained color reconstruction in the Face Branch; iii) compensatory overfitting in the Inside Mouth Branch, resulting in unnatural shading and misalignment.

\begin{table*}[!htb]
\setlength\tabcolsep{5pt}

\begin{center}

\resizebox{\linewidth}{!}{
\begin{tabular}{@{}c@{\hspace{6pt}}lccc|ccc|cc@{}} 
\toprule
\multicolumn{2}{l}{\multirow{2}{*}{Methods}}
 & \multicolumn{3}{c|}{Rendering Quality} & \multicolumn{3}{c|}{Motion Quality} & \multicolumn{2}{c}{Efficiency} \\ 
\multicolumn{2}{l}{\multirow{2}{*}{}}     & PSNR $\uparrow$ & LPIPS $\downarrow$ & SSIM $\uparrow$  & LMD $\downarrow$ & AUE-(L/U) $\downarrow$ & Sync-C $\uparrow$ & Time & FPS \\
    \midrule
    \multirow{4}{*}{\rotatebox[origin=c]{90}{NeRF}} & AD-NeRF~\cite{guo2021ad} & 30.07 & 0.1042 & 0.9689 & 2.998 & 1.01/0.97 & 6.053 & 18.7h & 0.11  \\
                                                  & RAD-NeRF~\cite{tang2022real} & 31.21 & 0.0643 & 0.9733 & 2.961 & 0.95/0.83 & 5.726 & 5.3h & 28.7  \\
                                                  & ER-NeRF~\cite{li2023efficient} & 31.66 & 0.0385 & 0.9732 & 2.745 & 0.76/0.64 & 6.633 & 2.1h & 31.2 \\
                                                  & SyncTalk~\cite{peng2024synctalk} & \underline{33.90} & \underline{0.0246} & \underline{0.9950} & \underline{2.685} & \underline{0.67}/\underline{0.32} & \underline{7.672}  & 2.0h & 52 \\
    \midrule
    \multirow{3}{*}{\rotatebox[origin=c]{90}{3DGS}} & GaussianTalker~\cite{cho2024gaussiantalker} & 31.75 & 0.0494 & 0.9942 & 2.805 & 0.83/0.70 & 6.37 & 3.2h  & 95 \\
                                                  & TalkingGaussian~\cite{li2025talkinggaussian}
                                                  & 32.04 & 0.0319 & 0.9947 & 2.714 & 0.70/\underline{0.32} & 6.284 & \textbf{0.5h}  & 108 \\
    \cmidrule(l){2-10} 
                                                  & M2DAO-Talker (Ours) &  \textbf{34.47} & \textbf{0.0229} & \textbf{0.9967} & \textbf{2.636} & \textbf{0.61}/\textbf{0.26} & \textbf{7.756} & \underline{0.6h} & \textbf{150}  \\
    \bottomrule
\end{tabular}}\vspace{-2mm}
\caption{Quantitative results of Self-Reconstruction. We emphasize the highest and second-highest results by bolding and underlining the corresponding values, respectively.}
\label{tab:e1}
\end{center}
\vspace{-6mm}
\end{table*}
\noindent\textbf{Alternating Optimization Strategy.} To address this, we implement a decoupled, two-branch approach. We firstly optimize the Face Branch while freezing the Inside Mouth Branch, allowing stable estimation of coarse facial geometry and appearance, especially around the lip region. Then we jointly fine-tune both branches: enhancing color/opacity of the mouth region and refining facial outputs, enabling collaborative correction of previous approximations while avoiding destructive interference. 
This alternating optimization improves convergence, harmonizes overlapping regions, and maintains motion consistency. As shown in \cref{fig:MS}, it yields smoother point cloud geometry and enhanced appearance at facial-oral boundaries.

Given deformation parameters ($\delta_{\text{f}}$, $\delta_{\text{m}}$), Gaussian primitive ($\theta_{\text{f}}$, $\theta_{\text{m}}$) and camera parameters, we synthesize the final image through:
\begin{equation}
\label{eq:I_fuse}
\hat{I}_{\text{fuse}} = \hat{I}_{\text{f}} \times A_{\text{f}} + (\hat{I}_{\text{m}} \times A_{\text{m}} + I_{\text{bg}} \times (1 - A_{\text{m}})) \times (1 - A_{\text{f}}),
\end{equation}
and supervise it using the following reconstruction loss:
\begin{equation}
\label{eq:L_fuse}
\mathcal{L}_{\text{fuse}} = \mathcal{L}_{\text{s}}(\hat{I}_{\text{fuse}}, I) + \gamma \mathcal{L}_{\text{LPIPS}}(\hat{I}_{\text{fuse}}, I).
\end{equation}

\section{Experiments}
\subsection{Experimental Settings}
\noindent\textbf{Dataset.} 
Following prior works~\cite{ye2023geneface, li2023efficient, guo2021ad, tang2022real}, we use seven publicly available video clips to train and evaluate both our M2DAO-Talker and baselines. 
Each video averages 7,066 frames at 25 FPS, focusing on a centered portrait, which are $512\times512$  (``Obama2'',``May'', ``Shaheen''and ``Lieu'') or $450\times450$ (``Obama'', ``Jae-in'', and ``Obama1'') resolution. 


\noindent\textbf{Comparison Baseline.} 
We provide comprehensive comparison and analysis against several state-of-the-art 2D generative models (IP-LAP~\cite{zhong2023identity}, TalkLip~\cite{wang2023seeing} and DINet~\cite{zhang2023dinet}), NeRF-based approaches (AD-NeRF~\cite{guo2021ad}, RADNeRF~\cite{tang2022real}, ER-NeRF~\cite{li2023efficient} and SyncTalk~\cite{peng2024synctalk}) and 3DGS-based methods (GaussianTalker~\cite{cho2024gaussiantalker} and TalkingGaussian~\cite{li2025talkinggaussian}).

\begin{table}[!htb]
\setlength\tabcolsep{5pt}
\begin{center}

\resizebox{\linewidth}{!}{
\begin{tabular}{@{}c@{\hspace{6pt}}lccc@{}} 
\toprule
\multicolumn{2}{l}{\multirow{2}{*}{Methods}}
& \multicolumn{3}{c}{Motion Quality}  \\ 
\multicolumn{2}{l}{\multirow{2}{*}{}}       & LMD $\downarrow$ & AUE-(L/U) $\downarrow$ & Sync-C $\uparrow$  \\
    \midrule
    \multirow{3}{*}{\rotatebox[origin=c]{90}{GAN}}
                                                  & IP-LAP~\cite{zhong2023identity}  & 3.266 & 1.01/- & 7.047  \\
                                                  & DINet~\cite{zhang2023dinet}   & 3.366 & 1.18/- & 7.645  \\
                                                  & TalkLip~\cite{wang2023seeing}   & 3.376 & 1.01/- & 6.536  \\
    \midrule
    ~    & M2DAO-Talker (Ours)  & \textbf{2.636} & \textbf{0.61}/\textbf{0.26} & \textbf{7.756}  \\
    \bottomrule
\end{tabular}}\vspace{-2mm}
\caption{Comparison results of Self-Reconstruction with GAN-based methods. The best  results are indicated in bold.}
\label{tab:e11}
\end{center}
\vspace{-6mm}
\end{table}

\noindent\textbf{Evaluation Metrics.}
For static image quality, we report Peak Signal-to-Noise Ratio (PSNR), Structural Similarity Index (SSIM)~\cite{wang2004image}, and Learned Perceptual Image Patch Similarity (LPIPS)~\cite{zhang2018unreasonable} to assess overall reconstruction accuracy, structural integrity, and high-frequency detail preservation, respectively.
For dynamic motion evaluation, we employ SyncNet~\cite{chung2017lip} to measure lip-sync quality via Landmark Distance (LMD), Confidence Score (Sync-C), and Error Distance (Sync-D). To further quantify facial expressiveness, we 
report Upper-face and Lower-face AU errors (AUE-U / AUE-L) to assess the precision of facial motion in different regions like TalkingGaussian~\cite{li2025talkinggaussian}.

\begin{table*}[]
\setlength\tabcolsep{5pt}
\begin{center}

\resizebox{\linewidth}{!}{
\begin{tabular}{lcccc|cccc}
\toprule
\multirow{3}{*}{Method}   & \multicolumn{4}{c}{``Shaheen" Audio} & \multicolumn{4}{c}{``Lieu" Audio} \\
             & \multicolumn{2}{c}{``Obama"} & \multicolumn{2}{c}{``May"} & \multicolumn{2}{c}{``Obama"} & \multicolumn{2}{c}{``May"} \\   \cmidrule(l){2-9} 

             & Sync-D $\downarrow$   & Sync-C $\uparrow$ & Sync-D $\downarrow$   & Sync-C $\uparrow$  & Sync-D $\downarrow$   & Sync-C $\uparrow$  & Sync-D $\downarrow$   & Sync-C $\uparrow$  \\

DINet~\cite{zhang2023dinet}          
                                        & 8.606        &\underline{7.481}        & \underline{8.201}         & \underline{7.295}               
                                        &  8.450        &  6.851        & 8.226         & 6.470          \\ 
IP-LAP~\cite{zhong2023identity}          
                                        & 9.190            & 6.135                  & 9.819             & 5.316      
                                        & 10.175       & 4.316              & 9.392             & 5.077       \\ 
TalkLip~\cite{wang2023seeing}          
                                        & 10.967           & 4.754                  & 9.553             & 5.488      
                                        &  11.648       & 3.459              & 11.679             & 3.151       \\ \midrule

RAD-NeRF~\cite{tang2022real}         
                                        & 8.633        & 7.166                  & 12.012       & 3.054                                  
                                        & 9.051        & 6.163                  & 12.044       & 2.449                             \\
ER-NeRF~\cite{li2023efficient}      
                                        & \textbf{8.240}         & \textbf{7.628}             & 9.775             & 5.529           
                                        & \textbf{8.103}         & \textbf{7.179}             & 10.017             & 4.782         \\ 

SyncTalk~\cite{peng2024synctalk}          &  8.600       & 7.099          &  8.903        & 6.350                   
                                        & 8.903       & 6.350         & \underline{7.508}         & \underline{7.780}                 \\

GaussianTalker~\cite{cho2024gaussiantalker}      
                                        & 9.415         & 6.686          & 8.926       & 6.576             
                                        & 10.171         & 5.441    & 10.943       & 4.198           \\ 
                                        
TalkingGaussian~\cite{li2025talkinggaussian}     
                                        & 10.596         & 4.543          & 11.450       & 3.179             
                                        & 9.702         & 5.746          & 9.849       & 5.039            \\ \midrule
\textbf{M2DAO-Talker (Ours)}     
                                        & \underline{8.486}  &  7.255     & \textbf{7.667}    & \textbf{8.111}     
                                        & \underline{8.371}         & \underline{6.916}          & \textbf{7.028}    & \textbf{8.333}            \\ \bottomrule
\end{tabular}
 }
\end{center}\vspace{-4mm}
\caption{Quantitative results of cross-domain audio-driven Lip Synchronization. }
\label{tab:e2}
\vspace{-4mm}
\end{table*}

\noindent\textbf{Implementation Details.} 
We adopt a two-phase training scheme per identity: 50K iterations of joint training for both branches, followed by 20K iterations of alternating optimization. The network is trained using Adam and AdamW with a learning rate of $5\times10^{-4}$. In the loss functions (\cref{eq:Lm_face,eq:Ls_face,eq:L_fuse}), we set $\lambda = 0.2$ and $\gamma = 0.5$. All experiments are conducted on NVIDIA RTX 3090 GPUs.

\subsection{Comparison with SOTA}

\noindent\textbf{Comparison Settings.}
In our quantitative evaluation, we assess the proposed method under two distinct conditions: the \emph{self-reconstruction setting} and the \emph{lip synchronization setting}. In the self-reconstruction setting, a 10-to-1 train-validation split is applied to each dataset to evaluate reconstruction fidelity. For the lip synchronization setting, we perform cross-domain evaluation using audio from unseen speakers (“Lieu” and “Shaheen”) to drive target identities (“Obama” and “May”), including cross-gender cases, to test generalization and synchronization robustness.

\begin{figure*}[!ht]
    \centering
    \includegraphics[width=\linewidth]{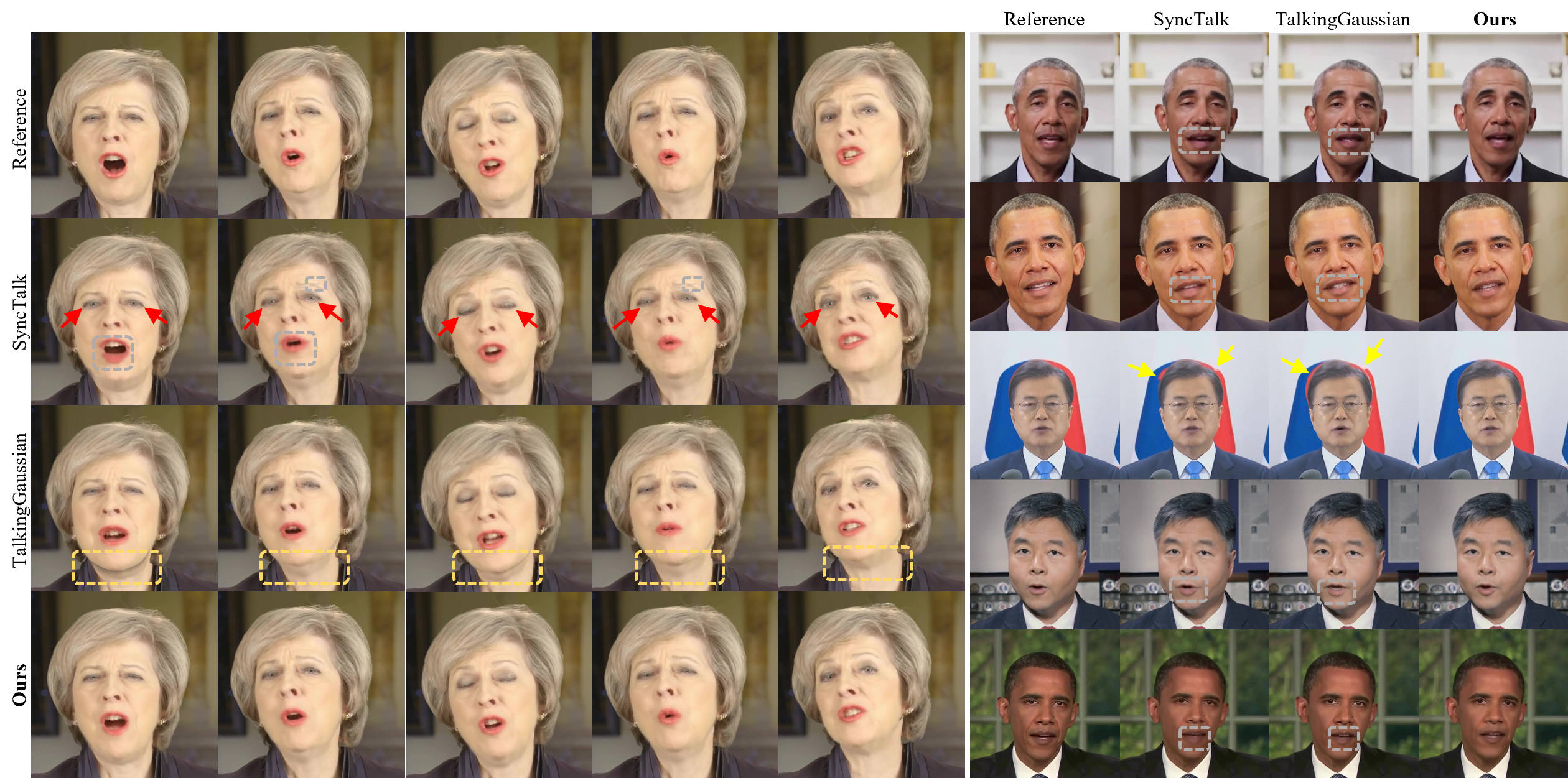}
    \vspace{-6mm}
    \caption{Qualitative results of Image Quality Comparison. Our method effectively eliminates edge blur and corrects facial distortions, blinking errors, and unnatural head-torso synthesis that were present in previous approaches. For improved visualization, please zoom in or refer to Supplementary Material.}
    \label{fig:IQ}
\vspace{-4mm}
\end{figure*}
\noindent\textbf{Quantitative Results.}
Our method surpasses existing approaches in most metrics (\cref{tab:e1,tab:e11}). M2DAO-Talker achieves the highest PSNR of \textbf{34.47} and lowest LPIPS of \textbf{0.0229}, outperforming SyncTalk (33.90) and TalkingGaussian (32.04) by leveraging improved preprocessing and region-specific deformation. In motion accuracy, it achieves an LMD of \textbf{2.636} and AUE-L of \textbf{0.61}, indicating effective motion disentanglement. Moreover, a Sync-C of \textbf{7.756} exceeds even specialized 2D lip-sync baselines, thanks to our alternating optimization strategy. In cross-domain lip synchronization (\cref{tab:e2}), our method demonstrates stable and consistent performance across identities and genders. Particularly in “Lieu” to “May”, it maintains accurate synchronization despite large speaker variation, highlighting its ability to generalize phoneme-to-motion mapping independently of identity. While ER-NeRF performs well in certain scenarios, it fluctuates significantly due to its direct audio-to-lip mapping, which lacks robustness to identity shifts. In contrast, M2DAO-Talker’s multi-granular motion modeling ensures reliable identity-consistent synthesis across diverse speaker conditions.

\noindent\textbf{Image Quality Comparison.} 
Figure \ref{fig:IQ} visually compares self-reconstruction results from SyncTalk, TalkingGaussian, and our method. Our approach achieves superior facial detail sharpness and cleaner head-background boundaries.
SyncTalk exhibits facial distortions (gray boxes/red arrows) due to inadequate motion field modeling, particularly misarticulating lip and eye regions. Our motion disentanglement resolves these limitations.
TalkingGaussian shows head-torso kinematic discontinuity (yellow box) and blending artifacts from poor segmentation (yellow arrows). Our Motion Consistency Constraint (MCC) eliminates discontinuity while hybrid segmentation ensures precise mask boundaries and artifact-free foreground-background fusion. 


\begin{figure}[!htb]
    \centering
    \includegraphics[width=\linewidth]{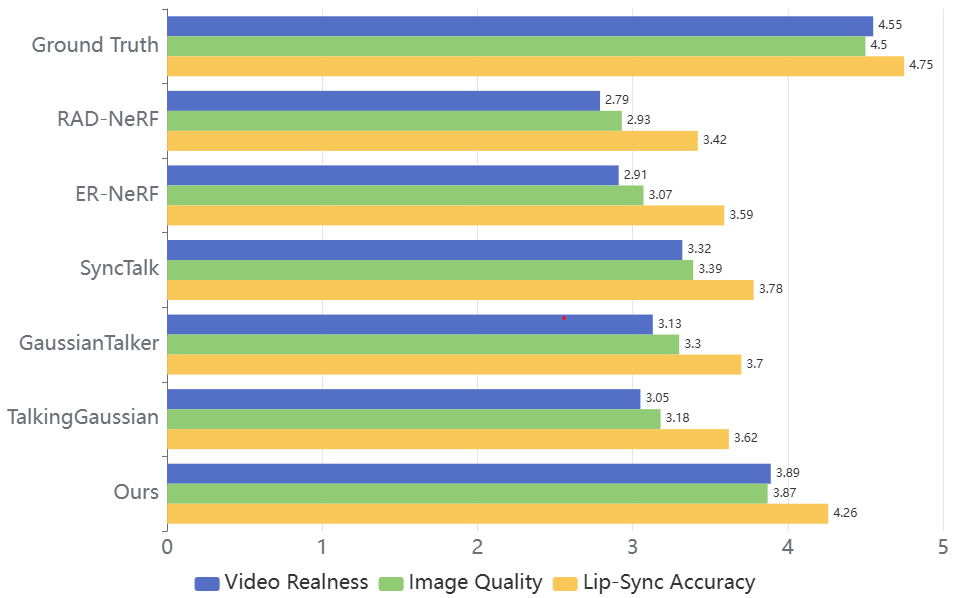}
    \vspace{-6mm}
    \caption{User Study. The rating is in the range of 1-5, higher denotes better performance.}
    \label{fig:US}
\vspace{-3mm}
\end{figure}

\noindent\textbf{User Study.}
To evaluate perceptual quality, we conduct a user study with 20 participants rating 49 half-minute videos from six models and Ground Truth. Each subject scores seven videos (1–5 scale) on the video realness, image quality and lip-sync accuracy. As shown in \cref{fig:US}, our M2DAO-Talker achieves the highest overall ratings across all criteria, reaching 87\% of GT scores on average  and the lowest lip-sync error (0.49). These results highlight the perceptual advantages of our motion disentanglement and alternating optimization design.

\subsection{Ablation Study}
\noindent\textbf{Ablation Setting.} To assess the contribution of each module, we conduct ablation studies on key components across our framework. For MRS and HPE, which involve specific backbone design choices, we evaluate by replacing them with widely adopted alternatives under controlled settings. Specifically, we substitute MRS with BiseNet~\cite{yu2018bisenet} and conduct experiments on datasets such as ``Jae-in'',``Shaheen'', and ``May'' where boundary ambiguity is prominent. For HPE, we remove the high flow error frames filter and replace the other methods' estimations with HPE. We evaluate the results on the “Obama”, “Obama1”, and “Obama2” sequences, which feature diverse head movements.
In contrast, we ablate AOS and MCC on all seven datasets to comprehensively assess the global impact. 

\begin{table}[!b]
\vspace{-4mm}
\setlength\tabcolsep{3pt}
\renewcommand{\arraystretch}{1.1}
\begin{center}

\resizebox{\linewidth}{!}{
\begin{tabular}{@{}l|cc|ccc@{}}
\toprule
            Backbone & MRS &  BN &                PSNR $\uparrow$ & Sync-C $\uparrow$ & LMD $\downarrow$   \\ 
                                    \midrule
\multirow{2}{*}{M2DAO-Talker (Ours)}        & \checkmark   &     & \textbf{35.041} & \textbf{7.965}  & \textbf{2.4575}  \\ 
     &      & \checkmark   & 34.400            & 7.956             & 2.4586 \\
\midrule
\multirow{2}{*}{TalkingGaussian~\cite{li2025talkinggaussian}}      &                      & \checkmark   & 31.519 & 5.971  & 2.5973 \\ 
 & \checkmark    & & \textbf{33.260} & \textbf{6.868}  & \textbf{2.5603} \\
\midrule
\multirow{2}{*}{SyncTalk~\cite{peng2024synctalk}}    &               & \checkmark   & 33.955 & 7.663  & \textbf{2.5374} \\ 
    & \checkmark  & & \textbf{34.355} & \textbf{7.819}  & 2.5733 \\
 \bottomrule
\end{tabular}
}
\end{center}\vspace{-4mm}
\caption{MRS ablation result.} 
\label{tab:e3}
\vspace{-2mm}
\end{table}

\begin{figure}[!t]
    \centering
    \includegraphics[width=1\linewidth]{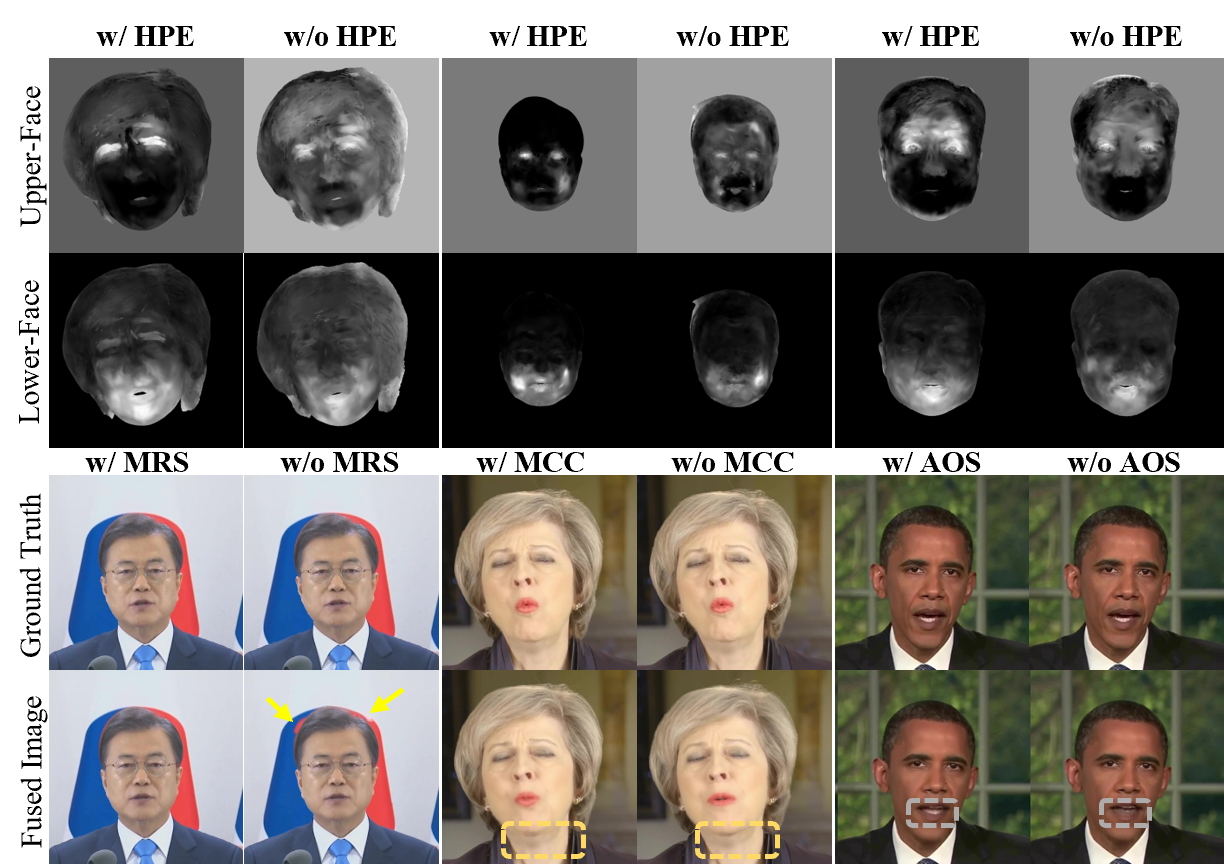}\vspace{-2mm}
\caption{Visualization of region attention map and reconstruction result in individual components ablation. } 
\vspace{-3mm}


    \label{fig:ablation1}
\end{figure}

\noindent\textbf{Ablation Results.}
Replacing the standard segmentation backbone with MRS yields consistent improvements (\cref{tab:e3}). These gains stem from MRS’s precise boundary localization, which reduces motion interference and enables more accurate region-specific deformation. Even SyncTalk, which lacks explicit deformation modeling, benefits from MRS, confirming its general utility for motion-aware synthesis.
Incorporating HPE also improves most metrics (\cref{tab:e6}), as it explicitly decouples rigid head rotation from facial dynamics, enabling smoother and more stable deformation learning, especially under large pose variations. With HPE, the attention maps of upper-face ($V_e$) and lower-face ($V_a$) focus more on the corresponding regions (\cref{fig:ablation1}), indicating that accurate decoupling of rigid head rotation enhances the learning of non-rigid facial motion.

Disabling MCC results in poor quality (\cref{tab:e7}) and inconsistency in motion, where face and torso move incoherently, leading to artifacts such as jagged contours and depth violations (\cref{fig:ablation1}). These results highlight MCC’s importance in enforcing biomechanical coherence by aligning facial motion with torso dynamics.
Replacing AOS with naive joint training causes color bleeding and perceptual degradation around the lips (\cref{tab:e7}). The lack of staged supervision leads to early overfitting in the Inside Mouth Branch, reducing lip detail fidelity and disrupting facial-oral continuity.



\begin{table}[!htb]
\setlength\tabcolsep{3pt}
\renewcommand{\arraystretch}{1.1}
\begin{center}
\vspace{-2mm}
\resizebox{\linewidth}{!}{
\begin{tabular}{@{}l|c|ccc@{}}
\toprule
               Backbone & HPE    & PSNR $\uparrow$ & Sync-C $\uparrow$ & LMD $\downarrow$  \\ 
                                    \midrule
\multirow{2}{*}{M2DAO-Talker (Ours)}                                                                     & \checkmark       & \textbf{33.746} &\textbf{ 7.372}  & 2.809  \\ 
&     & 33.648 & 7.338  & \textbf{2.799} \\
\midrule
\multirow{2}{*}{TalkingGaussian~\cite{li2025talkinggaussian}}                              &        & 32.384 & 6.836  & 2.825 \\ 
& \checkmark    & \textbf{33.215} & \textbf{6.859}  & \textbf{2.762} \\
\midrule
\multirow{2}{*}{SyncTalk~\cite{peng2024synctalk}}                                                     &    & 33.439 & \textbf{7.450} & 2.839 \\ 
& \checkmark     & \textbf{33.669} & 7.174  & \textbf{2.787} \\
 \bottomrule
\end{tabular}
}
\end{center} \vspace{-4mm}
\caption{HPE ablation result.} 
\label{tab:e6}\vspace{-4mm}

\end{table}

\begin{table}[!htb]
\setlength\tabcolsep{3pt}
\renewcommand{\arraystretch}{1.1}
\begin{center}

\resizebox{\linewidth}{!}{
\begin{tabular}{@{}l|cc|ccc@{}}
\toprule
Backbone & MCC & AOS         & PSNR $\uparrow$ & Sync-C $\uparrow$ & LMD $\downarrow$  \\ 
                                    \midrule
\multirow{3}{*}{M2DAO-Talker (Ours)}       & \checkmark & \checkmark                                                        & \textbf{34.475} & \textbf{7.757}  & 2.636 \\ 
        &  & \checkmark     & 34.123 & 7.688  & 2.624 \\

        & \checkmark &                   & 34.375 & 7.556  & \textbf{2.620} \\ 
 \bottomrule
\end{tabular}
}\vspace{-2mm}
\caption{MCC and AOS ablation result.} 
\vspace{-7mm}
\label{tab:e7}
\end{center}

\end{table}

\section{Conclusion}
We propose \textbf{M2DAO-Talker}, a novel audio-driven talking head framework that improves video stability and coherence through multi-granular motion disentanglement and alternating optimization. A robust 2D video preprocessing pipeline is developed to extract frame-wise deformation control, segment semantic motion regions, and estimate camera parameters. By decoupling head motion into rigid, facial, and oral components, and reinforcing coherence via a motion consistency constraint and alternating optimization, our approach effectively reduces timing jitter and artifacts, achieving high-quality, real-time synthesis.

\bibliography{aaai2026}
\newpage
\appendix

\twocolumn[
  \begin{center}
    {\LARGE \bfseries Appendix:M2DAO-Talker \par}
\vspace{2em}
  \end{center}
]

This appendix provides supplementary details to support the main paper. It includes related works, extended methodology, additional experiments, and further analysis. The appendix is structured as follows:

\begin{itemize}
\item \textbf{A. Related Works.}
We briefly review prior studies on audio-driven talking head synthesis and motion field modeling to contextualize our contributions.

\item \textbf{B. Video Preprocessing.}
We describe additional components omitted from the main text, including the \emph{Facial Action Encoder} and \emph{Audio-Visual Encoder}, which enhance motion disentanglement and region-specific modeling.

\item \textbf{C. More Ablation Study.}
We provide targeted ablation experiments that validate the individual contributions of the \emph{Facial Action Encoder} and \emph{Audio-Visual Encoder} to overall model performance.

\item \textbf{D. Additional Visualization.}
We present qualitative comparisons against SyncTalk and TalkingGaussian on high-resolution real-world data. We further analyze common failure modes to highlight our model's advantages in robustness and realism.

\item \textbf{E. Limitation.}
We discuss the limitations of our current pipeline, including challenges in audio encoding robustness, semantic mask generation, and the constraints of optical flow-based camera estimation.

\item \textbf{F. Future Work.}
We aim to extend our identity-specific modeling framework toward more generalizable talking head synthesis. Future directions include enabling arbitrary avatar identities to speak with diverse, transferable styles, supporting real-time customization and broader applicability in practical scenarios.

\end{itemize}

We hope these additional materials offer a deeper understanding of our approach and its broader implications.

\section{A. Related Works}
In this section, we briefly review the more recent advances in the field of talking head synthesis and motion field modeling.  

\subsection{Talking Head Synthesis}
Audio-driven talking head generation aims at generating videos with accurate lip movements and realistic facial animations based on input audio. 
Early methods, primarily based on 2D GANs~\cite{wang2023seeing, zhang2023dinet, zhong2023identity, guan2023stylesync, sun2022masked, prajwal2020lip}, have achieved considerable progress in terms of photorealism. 
However, 
only the lip regions are reconstructed~\cite{prajwal2020lip}, which neglects other facial movements, facial expressions in particular. 
Further, some efforts propose to learn the complete facial representation~\cite{lu2021live,wang2021audio2head}, but only display a fixed head posture due to the absence of a 3D geometric modeling. 

Compared to CNN-based technologies, NeRF~\cite{mildenhall2021nerf} shows impressive capability for continuous volumetric scene representation, which has garnered attention in employing for audio-driven talking head generation. 
For example, AD-NeRF~\cite{guo2021ad} 
proposes to project audio features to dynamic neural radiance fields, enabling portrait rendering by decoupling head and torso motion. 
To optimize modeling efficiency, researchers introduce the multi-resolution hash~\cite{tang2022real} to encode spatial and audio information independently while decomposing the 3D motion field into three-plane hash representation to minimize collisions~\cite{li2023efficient}. 
Meanwhile, SyncTalk~\cite{peng2024synctalk} emphasizes synchronization as a key challenge, and introduces an Audio-Visual Encoder for better audio-visual alignment. 

Despite significant advancements, NeRF-based methods learn implicit relationships among position, color, and density, which inevitably suffer from facial distortion and unnatural deformation due to the absence of geometrical constraints. 
Compared to NeRF, 3DGS utilizes the explicit geometry information to model scene, which shows faster and more efficient rendering while preserving fine details. 
More recently, 3DGS has been employed for audio-driven talking head generation~\cite{10707196,cho2024gaussiantalker,li2025talkinggaussian}. For example, GaussianTalker~\cite{cho2024gaussiantalker} proposes to learn mutual refinement of spatial-audio information, and characterizes motion field to improve facial fidelity and lip synchronization. 
Building on this, TalkingGaussian~\cite{li2025talkinggaussian} decouples oral movements and facial expressions to achieve finer grained motion representation, while using incremental sampling methods to improve compatibility between 3DGS and deformation. 

However, these approaches still face challenges, including edge ambiguities in semantic masks, incomplete decoupling of head motion and inconsistency between torso and facial motion. 
To address these issues, we introduce 2D segmentation priors to refine semantic parsing and head motion priors to optimize head pose estimation. Meanwhile, we learn the head-torso unified representation to maintain motion consistency, enhancing the quality of the generated video.

\subsection{Motion Field Modeling}  
Prior to deep learning, previous methods~\cite{horn1981determining, wedel2009structure,memin1998dense,brox2004high} focus on object-wise motion modeling, and estimate pixel-wise motion vectors using brightness constancy constraints. 
And then researchers introduce CNNs to regress motion vectors via an encoder-decoder structure~\cite{dosovitskiy2015flownet, teed2020raft, huang2022flowformer, sun2022skflow}. 
 
However, 2D motion field-based models struggle with precise motion representation due to a lack of scene information constraints, leading to unnatural deformations under occlusion or viewpoint changes. 
NeRF-based approaches consider both geometric shapes and motion states, and learn the relationships among position, color, and density for 3D reconstruction of complex scenes~\cite{park2021nerfies, pumarola2021d, park2021hypernerf, tretschk2021non}. 
For instance, D-NeRF~\cite{pumarola2021d} uses a deformation network to project motion field coordinates to a NeRF-based canonical space, while Nerfies~\cite{park2021nerfies} correlates the motion fields with latent codes to model intricate scenarios, such as human motion. 
While gaining impressive rendering quality, due to the lack of explicit geometric constraints, these methods produce unsatisfied results with obvious motion distortions in the case of rapid movements. 
By explicitly regulating spatial point variations through 
motion field representation, 3DGS effectively models temporal changes in dynamic scenes~\cite{wu20244d, duan20244d, hu2024gaussianavatar,yang2023real,yang2024deformable,wang2024shape,huang2024textit}.

However,  existing 3DGS-based approaches primarily derive motion dynamics function from temporal sequences, where deformation fields are conditioned solely on frame indices or timestamps. 
Previous works reveal that nonverbal facial actions introduce motion variances uncorrelated with phoneme timing and features alter articulation dynamics beyond temporal alignment. 
It indicates that audio-driven motions are governed with multiple factors where time serves merely as a carrier variable. 
To bridge this gap, we extract phoneme-aligned mouth movements features and person-agnostic facial expression features from 2D video as control conditions. 
And we establish a nonlinear mapping from control conditions to Gaussian's deformation, synthesizing motions congruent with both audio content and expressive context. 

\section{B. Video Preprocessing}
\noindent\textbf{Audio-Visual Encoder (AVE)}
Talking head generation relies on strong correlations between phonemes and orofacial kinematics. While existing audio encoders (\emph{e.g.}, automatic speech recognition (ASR)-focused models) prioritize semantic extraction of speech signals, they fail to capture fine-grained mouth movement features critical for articulatory dynamics. 
To address this, we adopt SyncTalk’s pre-trained audio-visual encoder~\cite{peng2024synctalk}, adversarially trained on the LRS2 dataset~\cite{afouras2018deep} with a lip-sync discriminator. 
This model extracts phoneme-aware mouth movement features $a$, establishing cross-modal alignment between speech signals and articulatory motion. These features are injected into the deformation field via attention-based modulation, controlling lip-related Gaussian primitive deformations.

\noindent\textbf{Facial Action Encoder (FAE)}
Facial expressions during speech production involve complex muscle activations that extend beyond phoneme articulation, including subtle movements such as squinting, eyebrow elevation, and frowning. Current approaches exhibit limitations in modeling these dynamics. For example, some efforts regulate basic blinking patterns~\cite{li2023efficient, tang2022real,9858334} but fail to capture nuanced facial muscle coordination. Blendshape-driven techniques~\cite{peng2024synctalk} enable broader expression control through 3D shape approximations but lack direct correspondence to biomechanical muscle activations. 
To bridge this gap, 
we propose a facial action encoder grounded in the Facial Action Coding System (FACS)~\cite{ekman1978facial}. Our framework parameterizes facial dynamics using six anatomically defined Action Units (AUs: 1, 4, 5, 6, 7, 45), selected for two key advantages. 

AUs directly map to specific muscle groups (\emph{e.g.}, AU4 for brow lowering, AU6 for cheek elevation) without inducing global facial distortions common to blendshape approaches.
In addition, these AUs demonstrate low correlation with phoneme-driven lip movements, enabling independent synchronization of emotional expressions and speech.
During training, we extract frame-wise facial expression features $e\in\mathbb{R}^6$ via OpenFace~\cite{baltrusaitis2018openface}, focusing on AUs governing brow, cheek, and eyelid kinematics. 
These features are integrated into the deformation field through attention-based modulation, enabling precise control of Gaussian primitives associated with upper-face dynamics.

\section{C. More Ablation Study}
\noindent\textbf{Audio-Visual Encoder.} 
To assess the effectiveness of AVE, we conduct ablation experiments on datasets with constrained  head rotations (``Lieu", ``Shaheen" and ``Jae-in"), where subtle articulatory dynamics are critical for accurate lip synchronization. 
Specifically, for AVE-equipped methods (our M2DAO-Talker approach and SyncTalk~\cite{peng2024synctalk}): we replace AVE with DeepSpeech~\cite{amodei2016deep} (DS) audio encoding. 
For DS-based methods (TalkingGaussian~\cite{li2025talkinggaussian}), we substitute DS with AVE encoding. We maintain the identical architectures otherwise for fair comparison, and evaluate their performance across PSNR, Sync-C, and LMD. 
Experiments in Table 5 demonstrate that AVE implementations outperform DS variants across all metrics, which displays AVE's advantage in capturing mouth movement dynamics over DS's semantic speech features. It is speculated that AVE's focus on articulatory features  enhances visual-articulatory alignment. 
Significantly, our M2DAO-Talker method maintains performance parity with SyncTalk and TalkingGaussian even when using suboptimal DS encoding, demonstrating obvious architectural resilience and superiority. 

\begin{table}[!ht]
\setlength\tabcolsep{3pt}
\renewcommand{\arraystretch}{1.1}
\begin{center}

\label{tab:e4}
\resizebox{\linewidth}{!}{
\begin{tabular}{@{}l|cc|ccc@{}}
\toprule
        Backbone & AVE & DS  & PSNR $\uparrow$ & Sync-C $\uparrow$ & LMD $\downarrow$  \\ 
                                    \midrule
\multirow{2}{*}{M2DAO-Talker (Ours)}                                                                          & \checkmark   &    & \textbf{35.932} &  \textbf{7.793} &  \textbf{2.423} \\ 
   &    &  \checkmark & 35.672 & 6.560  & 2.484 \\
\midrule
\multirow{2}{*}{TalkingGaussian~\cite{li2025talkinggaussian}}     &    &  \checkmark                                 & 32.687 & 6.442  & 2.484 \\ 
 & \checkmark   &     & \textbf{32.841} & \textbf{7.485} & \textbf{2.363} \\
\midrule
\multirow{2}{*}{SyncTalk~\cite{peng2024synctalk}}  & \checkmark   &                                               & \textbf{35.075} & \textbf{7.522}  & \textbf{2.503} \\ 
&    &  \checkmark     & 34.878 & 6.162  & 2.510 \\
 \bottomrule
\end{tabular}
}
\end{center}
\caption{Results of Audio-Visual Encoder Experiment. We compared the performance of the Audio-Visual Encoder (AVE) and Deepspeech (DS). The best  results are indicated in bold.}
\end{table}

\begin{table}[!hb]
\setlength\tabcolsep{3pt}
\renewcommand{\arraystretch}{1.1}
\begin{center}

\label{tab:e5}
\vspace{-1em}
\resizebox{\linewidth}{!}{
\begin{tabular}{@{}l|cc|ccc@{}}
\toprule
        Backbone & AU & BS  & PSNR $\uparrow$ & Sync-C $\uparrow$ & LMD $\downarrow$  \\ 
                                    \midrule
\multirow{2}{*}{M2DAO-Talker (Ours)}        & \checkmark   &                    & \textbf{34.475} & \textbf{7.757}  & \textbf{2.636} \\ 
&    &  \checkmark      & 34.333 & 7.714  & 2.672 \\
\midrule
\multirow{2}{*}{TalkingGaussian~\cite{li2025talkinggaussian}}   & \checkmark   &                     & \textbf{32.042} & 6.387  & 2.712 \\ 
&    &  \checkmark        & 31.938 & \textbf{6.397}  & 2.712 \\
\midrule
\multirow{2}{*}{SyncTalk~\cite{peng2024synctalk}}    &    &  \checkmark                          & 33.902 & 7.671  & 2.685 \\ 
& \checkmark   &      & \textbf{33.905} & \textbf{7.675}  & \textbf{2.660} \\
 \bottomrule
\end{tabular}
}
\end{center}
\caption{Results of Facial Action Encoder Experiments. We compared the performance of the Action Unit (AU) and Blendshape (BS). The best results are indicated in bold.}
\vspace{-3mm}
\end{table}

\noindent\textbf{Facial Action Encoder.} 
To evaluate the efficacy of FAE, we conduct cross-paradigm ablation experiments by substituting Action Units (AUs) and Blendshapes (BS) across seven datasets with varying expression complexity. 
For AU-based methods (our M2DAO-Talker approach and  TalkingGaussian~\cite{li2025talkinggaussian}), we replace AUs with BS. 
Conversely, for BS-based methods, we substitute BS with AUs for facial expression control. 
The performance comparison across PSNR, Sync-C, and LMD are shown in Table 6, revealing that AU-to-BS substitution causes consistent metric degradation on average (PSNR: $-$0.142 dB, LMD: $+$0.036).
Overall, AUs outperforms BS in almost scenarios, particularly in complex and diverse expression scenarios (\emph{e.g.}, ``May":$+$0.61 dB, ``Obama1" :$+$0.29 dB and ``Obama2" :$+$0.19 dB). This validates AUs' superiority in modeling fine-grained muscle activations critical for subtle expressions.

\begin{figure*}[!h]
    \centering
    \includegraphics[width=\linewidth]{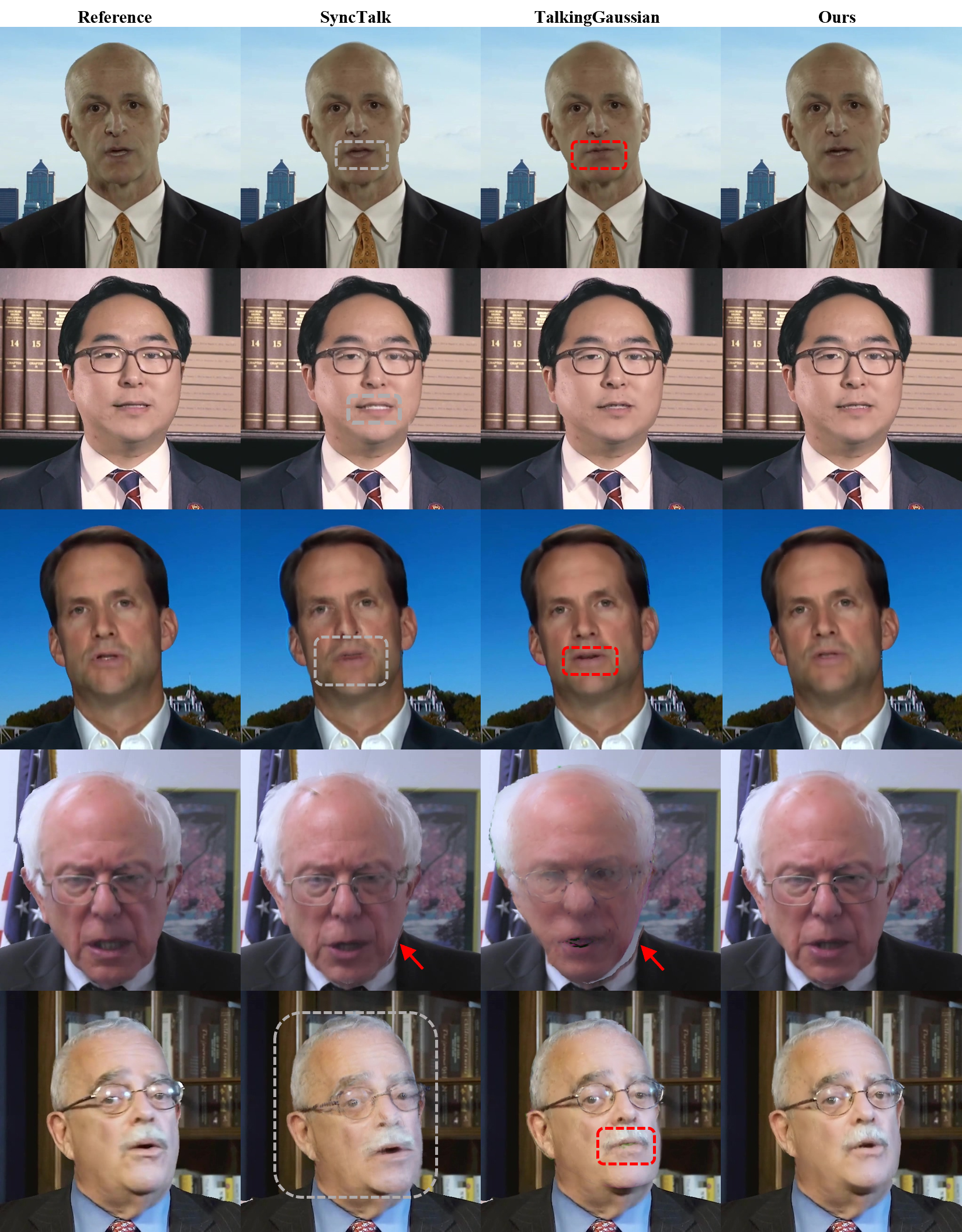}
    \vspace{-6mm}
    \caption{High-definition Comparison. We compare the generation results of different methods across five complex scenarios. Overall, our method achieves the best subjective visual quality.}
    \label{fig:vis}
\vspace{-4mm}
\end{figure*}

\section{D. Additional Visualization}
To facilitate a more comprehensive comparison with SyncTalk and TalkingGaussian, we select five speech videos exceeding four minutes in duration from the High-Definition Talking Face (HDTF) dataset~\cite{zhang2021flow}. Unlike the datasets used in the main paper, these videos feature substantial torso motion and large-scale head rotations during speech, thereby presenting challenging real-world conditions that stress model generalization and robustness. We adopt the same self-reconstruction setup as in earlier experiments, and visualize the results in \cref{fig:vis}. The following analysis highlights qualitative differences across the three methods.

\noindent\textbf{SyncTalk.}
Although both our method and SyncTalk employ AVE as the audio encoder, SyncTalk lacks an explicit motion field formulation. Instead, it directly predicts point cloud attributes from control signals, which leads to facial distortions under rapid articulatory motion—manifesting as blurred lips and degraded intra-oral regions (gray boxes). Moreover, fast head rotations often induce global facial deformations and motion blur. In contrast, our method decouples motion via a static Gaussian field combined with a dynamic motion field that predicts deformation parameters from control inputs, effectively alleviating such artifacts.
In addition, while both methods utilize optical flow for camera pose estimation, SyncTalk suffers from inaccurate pose predictions, resulting in face-torso misalignment and visible segmentation boundaries (red arrows). Our method addresses this limitation by improving the semantic segmentation network to produce more accurate facial masks, thereby enhancing the quality of optical flow supervision. A subsequent outlier removal step further filters unreliable frames during camera optimization. This joint enhancement of motion priors and pose stability ensures accurate head projection and clean spatial alignment between the face and torso.

\noindent\textbf{TalkingGaussian.}
While both TalkingGaussian and our method adopt a dual-branch architecture to separately model facial and intra-oral regions, the former relies heavily on the accuracy of facial segmentation masks. Additionally, TalkingGaussian uses DeepSpeech as its audio encoder, which is less effective in capturing fine-grained motion cues from speech, resulting in inaccurate lip motion and blurry intra-oral rendering (red boxes). In contrast, we adopt AVE for its stronger motion awareness and implement an alternating optimization strategy between the two branches, which reduces dependency on mask quality and enables more precise lip synchronization.
TalkingGaussian estimates camera parameters using a per-frame 3DMM fitting strategy, but lacks temporal consistency constraints. Furthermore, it computes losses across all 3DMM landmarks indiscriminately, failing to distinguish between rigid (e.g., upper face contours) and non-rigid (e.g., lip landmarks) components. As a result, the estimated camera poses are often biased by non-rigid deformations, leading to misaligned facial projections and noticeable disjunctions with the torso (red arrows), which ultimately degrades rendering fidelity.
In contrast, our method introduces an optical flow–based estimation of rigid facial keypoints and jointly optimizes camera parameters across all frames, achieving temporally consistent and accurate head pose tracking. Additionally, we enforce a motion consistency constraint (MCC) to encourage coherent deformation between the face and torso, producing more natural and anatomically plausible head movements.

\section{E. Limitation}
Our preprocessing pipeline has several limitations. First, AVE may cause lip jitter or incomplete mouth closure on some datasets, likely due to suboptimal pretraining. Second, common facial segmentation models generate binary masks, leading to jagged boundaries; using alpha-aware masks could improve edge smoothness. Lastly, our optical flow–based camera estimation fails on edited videos with temporal discontinuities, restricting its use to continuous sequences.

\section{F. Future Work}
While current 3D talking head methods achieve high realism by training on identity-specific data, they lack flexibility for real-world applications such as real-time avatar customization or cross-identity speaking style transfer. In future work, we aim to generalize our identity-specific modeling approach to support arbitrary avatars driven by diverse speaking styles, improving the scalability and adaptability of digital human generation.

\end{document}